\documentclass{article}

\usepackage{microtype}
\usepackage{graphicx}
\usepackage{subfigure}
\usepackage{booktabs}
\usepackage{array}
\usepackage{tabularx}
\usepackage{caption}

\usepackage{hyperref}

\usepackage[accepted]{icml2025}

\usepackage{amsmath}
\usepackage{amssymb}
\usepackage{mathtools}
\usepackage{amsthm}

\usepackage[capitalize,noabbrev]{cleveref}

\theoremstyle{plain}

\theoremstyle{definition}

\theoremstyle{remark}

\usepackage[textsize=tiny]{todonotes}

\icmltitlerunning{Submission and Formatting Instructions for ICML 2025}

\begin{document}

\twocolumn[
\icmltitle{DiagNet: Detecting Objects using Diagonal Constraints on Adjacency Matrix of Graph Neural Network}



\icmlsetsymbol{equal}{*}

\begin{icmlauthorlist}
\icmlauthor{Chong Hyun Lee}{jej}
\icmlauthor{Kibae Lee}{jej}
\end{icmlauthorlist}

\icmlaffiliation{jej}{Department of Ocean System Engineering, Jeju National University}

\icmlcorrespondingauthor{C. H. Lee}{chonglee@jejunu.ac.kr}

\icmlkeywords{Machine Learning, ICML}

\vskip 0.3in
]
\printAffiliationsAndNotice{}



\begin{abstract}

We propose DaigNet, a new approach to object detection with which we can detect an object bounding box using diagonal constraints on adjacency matrix of a graph convolutional network (GCN). We propose two diagonalization algorithms based on hard and soft constraints on adjacency matrix and two loss functions using diagonal constraint and complementary constraint. The DaigNet eliminates the need for designing a set of anchor boxes commonly used. To prove feasibility of our novel detector, we adopt detection head in YOLO models. Experiments show that the DiagNet achieves 7.5 \% higher mAP$^{50}$ on Pascal VOC than YOLOv1. The DiagNet also shows 5.1 \% higher mAP on MS COCO than YOLOv3u, 3.7 \% higher mAP than YOLOv5u, and 2.9 \% higher mAP than YOLOv8.

\end{abstract}

\section{Introduction}

Object detection models based on convolutional neural network (CNN) have gained attention because of their superior performance. These models are generally categorized into two-stage and one-stage object detectors. Two-stage object detectors such as Region-based CNN (R-CNN) generate a set of candidate boxes using region proposal algorithms and refine the results through classification and post-processing \cite{ren2016faster, he2017mask}. In contrast, one-stage object detectors such as various YOLO models, perform detection in a single step. The one-stage detector usually provides faster processing speeds but results in slightly lower detection performance in general than those of two-stage detectors \cite{redmon2016you, redmon2017yolo9000}.

To address the performance issues of one-stage object detectors, training methods using multiple anchor boxes have been proposed \cite{8237586, redmon2018yolov3, bochkovskiy2020yolov4}. These methods distribute anchor boxes of various sizes across input image and perform detection on each box to estimate bounding boxes. This approach, however increases training complexity and leads to data imbalance as only a few anchor boxes contain objects. To address this issue, various data augmentation techniques have been introduced to improve the generalization performance of models \cite{yolov5, yolov8_ultralytics}. Nevertheless, the performance heavily depends on number and size of anchor box. The lack of proper data augmentation also can lead to significant degradation in training performance. Furthermore, existing one-stage detection models struggle to accurately predict bounding boxes because objects are usually confined to a specific region within a box. In particular, when multiple bounding boxes are overlapped, detection performance is limited without employing diverse combinations of anchor boxes.

To overcome these limitations, we introduce DiagNet, a new one-stage approach based on graph convolutional network (GCN). We use a single GCN as a neck and use the GCN output feature matrix and adjacency matrix for generating a diagonalized map, which is a new feature map and fed into detection head. To generate the diagonalized map, we propose four algorithms based on diagonal constraints on adjacency matrix of the GCN. Two of the proposed algorithms are based on element of diagonal adjacency matrix and the other two are based on loss function of single or complementary diagonal matrix. To show the detection performance of the DiagNet, we adopt detection head in YOLO models and show the DiagNet improves mAP performance of existing YOLO models.

The main contributions of this paper are as follows:
\begin{itemize}
\item We propose a new GCN based object detection model which is composed of diagonalization algorithm, node embedding and edge prediction.
\item We propose two diagonalization algorithms based on hard and soft constraints on adjacency matrix.
\item We propose two loss functions using diagonal constraint and complementary constraint.
\item We verify detection performance improvement of the DiagNet by adopting detection head of the YOLO.
\end{itemize}

\section{Related works}
\subsection{Two-stage object detectors}
The two-stage object detector generates multiple regions of interest (ROIs) using region proposal algorithm, followed by object detection within those regions. Prominent two-stage object detectors include R-CNN \cite{girshick2014rich}, Fast R-CNN \cite{Girshick_2015_ICCV}, and Faster R-CNN \cite{ren2016faster}.

The R-CNN employs a selective search algorithm to extract multiple ROIs within an image, and detect objects in each ROI using a CNN and support vector machine (SVM) \cite{girshick2014rich}. Although R-CNN achieves superior detection performance, the individual processing of each ROI leads to considerable computational time. The Fast R-CNN generates a feature map of the image using a CNN and performs detection by extracting ROIs from the feature map \cite{Girshick_2015_ICCV}. This approach eliminates the feature extraction process for multiple ROIs in R-CNN, significantly reducing computational time.

The Faster R-CNN replaces the individual operated selective search algorithm in R-CNN and Fast R-CNN with learnable region proposal network (RPN), significantly reducing computational time \cite{ren2016faster}. Unlike the selective search algorithm, the RPN takes the feature map as input and directly extracts ROIs, enabling real-time detection.

\subsection{One-stage object detectors}
Unlike two-stage object detectors, one-stage detectors directly detect objects from the input image without generating ROIs. This approach enables significantly faster detection than two-stage object detectors. The YOLO models are well-known representative detectors \cite{redmon2016you, redmon2017yolo9000, redmon2018yolov3, bochkovskiy2020yolov4, yolov5, yolov8_ultralytics}.

The YOLOv1 partitions the image into multiple grids and detects objects by predicting bounding boxes for each grid cell \cite{redmon2016you}. Even the YOLOv1 enables fast object detection through single processing step, it struggles with detecting small objects and multiple overlapped objects. The low resolution of grid cells also results in low detection performance.

The YOLOv3 enhances detection performance over previous YOLO models by utilizing multi-scale feature maps and a residual network \cite{redmon2018yolov3}. To detect objects of varying sizes, the YOLOv3 employs a feature pyramid network, which can efficiently process multi-scale feature maps. The YOLOv3 is widely adopted in various applications because of its fast processing and superior detection performance over two-stage object detectors such as those in the R-CNN series \cite{choi2019gaussian, lawal2021tomato, shen2023improved}.

The YOLOv5 achieves even faster detection by utilizing a network architecture with fewer parameters than previous YOLO models \cite{yolov5}. To maintain high detection performance with a smaller network, the YOLOv5 applies various data augmentation techniques such as Mosaic augmentation, Mixup, and CutOut.

The YOLOv8 is an advanced model integrating and improving strengths of previous YOLO models \cite{yolov8_ultralytics}. It adopts a transformer architecture to efficiently extract multi-scale feature maps. The YOLOv8 also automates the anchor box generation process and optimizes the network modules for objection detection and bounding box prediction. These improvements contribute to more stable training and enhanced detection performance. After optimized modules in the YOLOv8 being applied to YOLOv3 and YOLOv5, further detection performance is obtained \cite{yolov5}.

One-stage object detectors such as RetinaNet \cite{8237586} and FCOS \cite{9010746} are reported along with YOLO models. The RetinaNet utilizes focal loss which makes it robust to data imbalance in training with numerous anchor boxes \cite{8237586}. The FCOS eliminates the need for anchor boxes by treating each pixel as the center of an object during training \cite{9010746}. Even the FOCS performs object detection without anchor boxes, its pixel-level processing during training increases computational complexity.

\section{Detection algorithm}
The overall framework of the proposed DiagNet is illustrated in Figure 1. The proposed one-stage object detector is composed of backbone network extracting input feature, neck network processing the extracted features and detection head detecting object and estimating bounding box of the object. The proposed DiagNet based on GCN is used as neck network and its output is transferred to detection head. The GCN is a type of neural network explicitly designed for processing graph-structured data \cite{scarselli2008graph} and has graph convolutional layer as its core component \cite{kipf2017semisupervised}. The graph convolutional layer enables localized spectral filtering to capture relationships between nodes characterizing the image patches. These neighborhood relationships are measured using the cosine similarity-based correlation.

\begin{figure}[ht]
\vskip 0.0in
\begin{center}
\centerline{\includegraphics[width=\columnwidth]{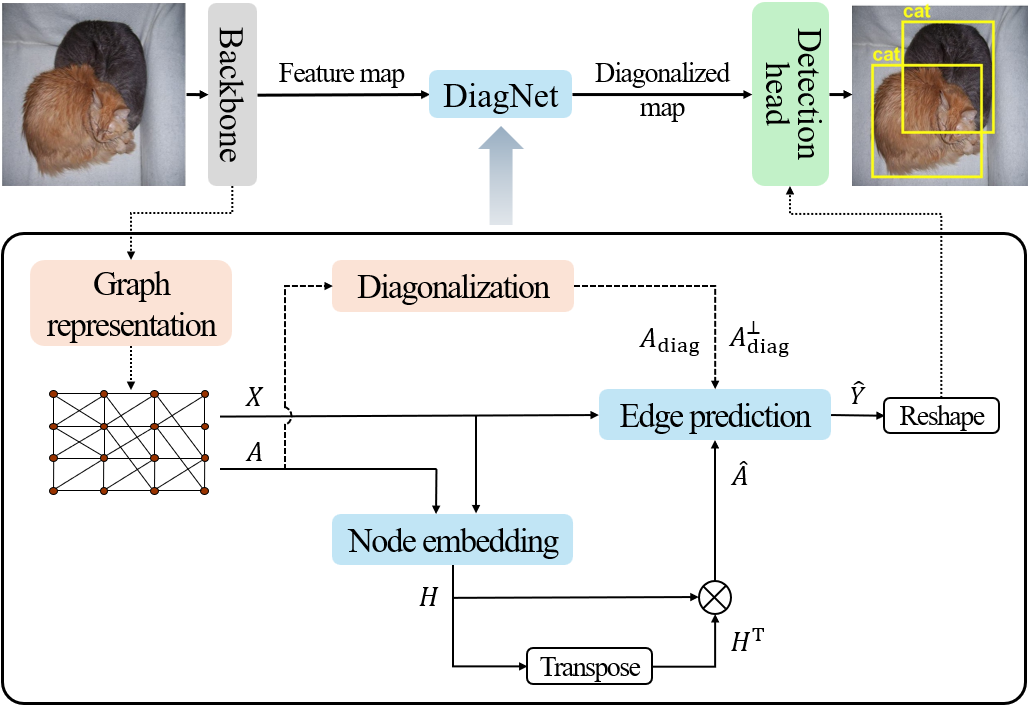}}
\caption{Overall framework of the DigNet.}
\label{icml-historical}
\end{center}
\vskip -0.2in
\end{figure}

\begin{figure}[ht]
\vskip 0.0in
\begin{center}
\centerline{\includegraphics[width=\columnwidth]{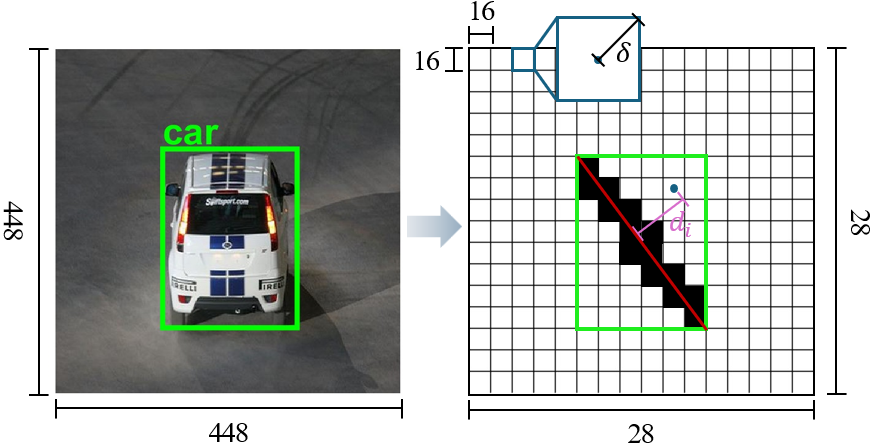}}
\caption{The diagonalization process for image of $h_{\text{in}}=448$.}
\label{icml-historical}
\end{center}
\vskip -0.2in
\end{figure}

The first stage of the DiagNet is to transform input feature to graph $G(X,A)$, where $X$ and $A$ are node matrix and adjacency matrix, respectively. By using the GCN, node embedding and edge prediction are performed simultaneously. For the edge prediction, we propose new diagonalization process which generates a diagonal matrix corresponding to object bounding box. For training the DiagNet, we define new loss function handing the generated diagonal matrix. The final stage is to transfer the DiagNet output to detection head based on YOLO model.

Suppose the output of the Backbone be a tensor size of $h \times h \times c$, where $h$ and $c$ are height and channel numbers, respectively. Then, this tensor is transformed to graph $G(X, A)$, where $X \in \mathbb{R}^{N \times L}$ and $A \in \mathbb{R}^{N \times N}$ are node and adjacency matrices using node number $N = h^2$ and feature dimension $L = c$. The $(i, j)$ component of $A$, $a_{ij}$ can be estimated as follows:

\begin{equation}
a_{ij} = \max \left( \frac{\mathbf{x}_i \mathbf{x}_j^T}{\|\mathbf{x}_i\| \|\mathbf{x}_j\|}, 0 \right), \quad \forall i,j
\end{equation}

where $\|\cdot\|$ represents vector norm and $\mathbf{x}_i$ and $\mathbf{x}_j$ are feature vectors of the $i$th node and $j$th node of $X$, respectively. Training procedure of the DiagNet can be described as following:

1) Node embedding:

\begin{equation}
H = \tanh \left( \tilde{A} X W_{\text{emb}} \right)
\end{equation}

where $H \in \mathbb{R}^{N \times L'}$ is a node matrix with reduced feature size $L'$, and $W_{\text{emb}} \in \mathbb{R}^{L \times L'}$ is a learnable weight matrix. The $\tilde{A}$ represents an adjacency matrix normalized degree matrix.

2) Edge prediction:

\begin{equation}
\hat{Y} = \tanh \left( X^T \hat{A} W_{\text{pred}} \right)
\end{equation}

where, $\hat{A} = H H^T$ is an estimated adjacency matrix and $W_{\text{pred}} \in \mathbb{R}^{N \times N}$ is a weight matrix for edge prediction. The basic DiagNet is trained by using a loss function $L_{\text{min}}$ defined as follows:

\begin{equation}
L_{\text{min}} = \| \hat{Y} - X^T A_{\text{diag}} \|
\end{equation}

where $A_{\text{diag}}$ represents a diagonal matrix corresponding to object bounding box. The $(i, j)$ element of $A_{\text{diag}}$ is estimated by using the following:

\begin{equation}
A_{\text{diag}}(i,j) =
\begin{cases} 
1, & d_i \leq \delta \text{ and } d_j \leq \delta \\
0, & \text{otherwise}
\end{cases}, \quad \forall i,j
\end{equation}

where $\delta = \frac{h_{\text{in}}}{2h} \sqrt{2}$ is a threshold to be determined. The $h_{\text{in}}$ represents the height of the input square image. If the size of the input image is $448 \times 448$, then $h_{\text{in}} = 448$, $h = 28$, and $\delta = 8\sqrt{2}$ as shown in Figure 2. Here we assume that the input image is divided into square patches, and each node represents the feature vector of the corresponding patch. The center of each patch is defined as the node’s coordinate in the graph. The $\delta$ corresponds to half of the diagonal length and then is used to determine the relationship between diagonal line and the node of the bounding box. The $d_i$ or $d_j$ represents the minimum distance from $i$th or $j$th node to the diagonal line. If the distance $d_i$ is less than or equal to $\delta$, the $i$th patch is considered to intersect with the diagonal line. Furthermore, when both $i$th and $j$th patches intersect with the diagonal line, the $i$th and $j$th nodes are connected. This diagonalization process is illustrated in Figure 2.

Similarly, we define $A_{\text{diag}}^{\perp}$ as a complementary matrix to $A_{\text{diag}}$. The $(i, j)$ element of $A_{\text{diag}}^{\perp}$ can be estimated by using the following:

\begin{equation}
A_{\text{diag}}^{\perp} (i,j) =
\begin{cases} 
0, & d_i \leq \delta \text{ or } d_j \leq \delta \\
1, & \text{otherwise}
\end{cases}, \quad \forall i,j
\end{equation}

Examples of $A_{\text{diag}}$ and $A_{\text{diag}}^{\perp}$ normalized by node-wise degree values, are shown in Figure 3 (b) and (c), respectively.

\begin{figure}[ht]
\vskip 0.0in
\begin{center}
\centerline{\includegraphics[width=\columnwidth]{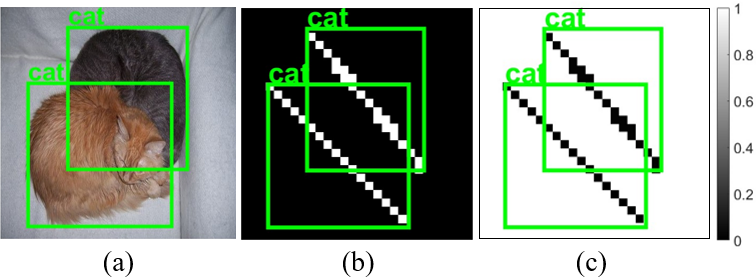}}
\caption{(a) Image with true bounding box, (b) $A_{\text{diag}}$ and (c) $A_{\text{diag}}^{\perp}$ normalized by node-wise degree values.}
\label{icml-historical}
\end{center}
\vskip -0.2in
\end{figure}

\begin{figure}[ht]
\vskip 0.1in
\begin{center}
\centerline{\includegraphics[width=\columnwidth]{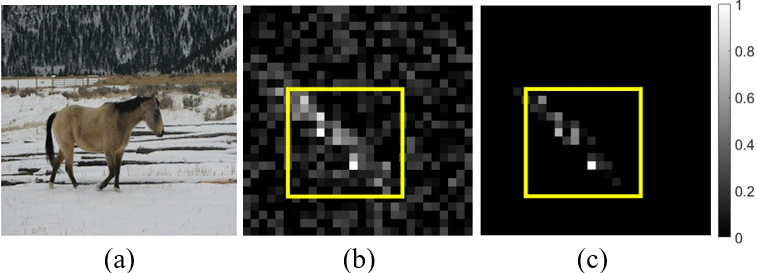}}
\caption{(a) Image, diagonal maps visualized via node-wise norms when (b) $L_{\text{min}}$ and (c) $L_{\text{comp}}$ are used with (5) and (6).}
\label{icml-historical}
\end{center}
\vskip -0.4in
\end{figure}

To improve detection performance further, we define a loss function $L_{\text{comp}}$ using a complementary constraint as follows:

\begin{equation}
L_{\text{comp}} = \frac{\| \hat{Y} - X^T A_{\text{diag}} \|}{\| \hat{Y} - X^T A_{\text{diag}}^{\perp} \|}
\end{equation}

The diagonalized maps obtained from DiagNet output $\hat{Y}$ after node-wise normalization when $\hat{Y}$ is trained by $L_{\text{min}}$ and $L_{\text{comp}}$, are shown in Figure 4 (b) and (c). As shown in the experimental results in the next section, we obtain improved performance when we use $L_{\text{comp}}$ in learning process.

The two loss functions (4) and (7) use hard constraint, i.e. all $A_{\text{diag}}(i,j)$ and $A_{\text{diag}}^{\perp}(i,j)$ are either 0 or 1. This constraint may result in limited convergence or detection property. To release the constraint, we use a soft constraint so that $A_{\text{diag}}(i,j)$ and $A_{\text{diag}}^{\perp}(i,j)$ can be real number between 0 and 1. Let $\phi_i$ be the soft minimum distance from $i$th node to the diagonal line and the $\phi_i$ be defined as follows:

\begin{equation}
\phi_i = \exp \left( \frac{-d_i}{2\sigma^2} \right), \quad i \in \{1,2,\dots,N\}
\end{equation}

where $\sigma = \frac{\alpha \delta}{\sqrt{2}}$ is a constant to be determined by relaxation parameter $\alpha$. Then, $A_{\text{diag}}(i,j)$ and $A_{\text{diag}}^{\perp}(i,j)$ can be real numbers between 0 and 1 by equations following:

\begin{equation}
A_{\text{diag}}(i,j) = \phi_i \phi_j, \quad \forall i,j
\end{equation}
\begin{equation}
A_{\text{diag}}^{\perp}(i,j) = (1 - \phi_i)(1 - \phi_j), \quad \forall i,j
\end{equation}

\begin{figure}[ht]
\vskip 0.0in
\begin{center}
\centerline{\includegraphics[width=\columnwidth]{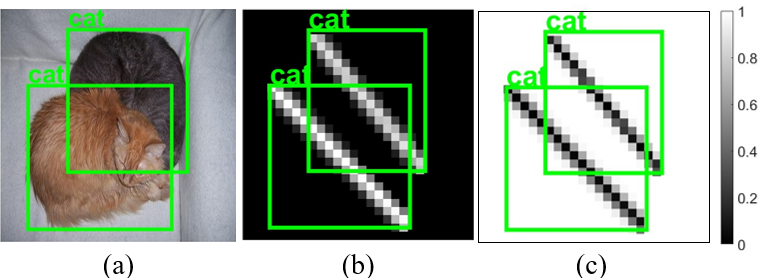}}
\caption{(a) Image with true bounding box, (b) $A_{\text{diag}}$ and (c) $A_{\text{diag}}^{\perp}$ normalized node-wise degree values and $\alpha = 1$ in (8).}
\label{icml-historical}
\end{center}
\vskip -0.2in
\end{figure}

\begin{figure}[ht]
\vskip -0.3in
\begin{center}
\centerline{\includegraphics[width=\columnwidth]{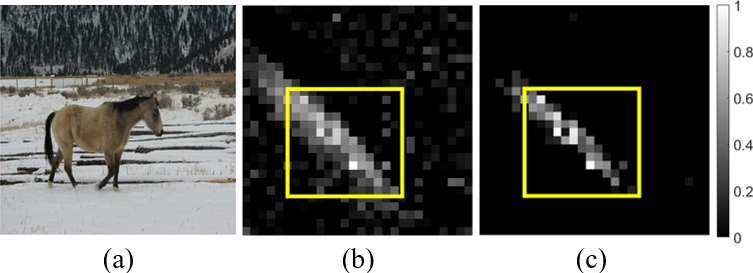}}
\caption{(a) Image, diagonal maps visualized via node-wise norms when (b) $L_{\text{min}}$ and (c) $L_{\text{comp}}$ are used with (9) and (10).}
\label{icml-historical}
\end{center}
\vskip -0.4in
\end{figure}

One example obtained by using (9) and (10) is shown in Figure 5 which clearly illustrates broader diagonal line as expected. The diagonalized maps trained by $L_{\text{min}}$ and $L_{\text{comp}}$ using (9) and (10) are shown in Figure 6 which clearly illustrate broader diagonal line than the results shown in Figure 4. By using $L_{\text{min}}$ and $L_{\text{comp}}$ with $\phi_i$ in (8), we can obtain best detection performance as in the following section.

\section{Experimental results}
\subsection{Dataset}

\begin{table*}[t]
\caption{Detection performance on Pascal VOC dataset.}
\label{tab:pascal-voc}
\vskip 0.15in
\centering
\begin{small}
\begin{sc}
\begin{tabular}{lcccccc}
\toprule
\textbf{Model} & \textbf{Backbone} & \textbf{Neck} & \textbf{Detection head} & \textbf{Diagonal loss} & \textbf{\#. Para.} & \textbf{mAP$^{50}$} \\
\midrule
YOLOv1~\cite{redmon2016you} & ResNet50 & CNN & FC layers & - & 281M & 50.7 \\
\midrule
DiagNet (hard) & ResNet50 & DiagNet & FC layers & $\mathcal{L}_{\text{min}}$ & 237M & 52.4 \\
 & & & & $\mathcal{L}_{\text{comp}}$ & 237M & 54.8 \\
\midrule
DiagNet (soft) & ResNet50 & DiagNet & FC layers & $\mathcal{L}_{\text{min}}$ & 237M & 54.3 \\
 & & & & $\mathcal{L}_{\text{comp}}$ & 237M & 58.2 \\
\bottomrule
\end{tabular}
\end{sc}
\end{small}
\vskip -0.1in
\end{table*}

\begin{figure*}[t]
\vskip 0.2in
\centering
\includegraphics[width=0.98\textwidth]{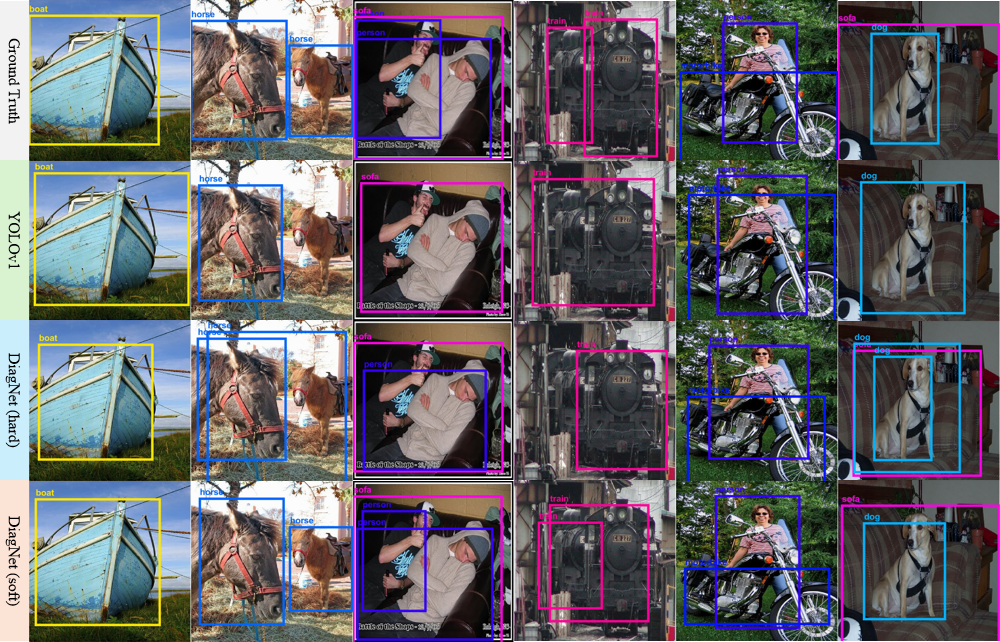}
\caption{Comparisons of detection results on Pascal VOC dataset.}
\label{icml-historical}
\vskip -0.2in
\end{figure*}

\begin{table*}[t]
\caption{Detection performance on MS COCO dataset.}
\label{tab:coco}
\vskip 0.2in
\centering
\begin{small}
\begin{sc}
\renewcommand{\arraystretch}{1.2}
\setlength{\tabcolsep}{3pt}
\begin{tabular}{p{4.4cm} p{1.8cm} p{1.4cm} p{2.0cm} p{1.4cm} c c c c}
\toprule
\textbf{Model} & \textbf{Backbone} & \textbf{Neck} & \textbf{Detection head} & \textbf{Diagonal loss} & \textbf{\#. Para.} & \textbf{mAP} & \textbf{mAP$^{50}$} & \textbf{mAP$^{75}$} \\
\midrule
\multicolumn{9}{l}{\textbf{Two-stage detectors}} \\
Faster R-CNN~\cite{ren2016faster} & ResNet50 & - & - & - & 41M & 36.9 & 58.8 & 39.8 \\
Mask R-CNN~\cite{he2017mask} & ResNet50 & - & - & - & 44M & 37.9 & 59.2 & 41.1 \\
\midrule
\multicolumn{9}{l}{\textbf{One-stage detectors}} \\
RetinaNet~\cite{8237586} & ResNet50 & - & - & - & 34M & 36.4 & 55.7 & 38.2 \\
FCOS~\cite{9010746} & ResNet50 & - & - & - & 32M & 39.1 & 58.2 & 42.0 \\
CornerNet~\cite{law2018cornernet} & Hourglass & Corner pooling layer & Multi-CNN & - & - & 42.2 & 57.8 & 45.2 \\
YOLOv3u~\cite{redmon2018yolov3, yolov5} & DarkNet53 & FPN & Ultralytics head & - & 103M & 48.0 & 64.7 & 51.9 \\
YOLOv5u~\cite{yolov5} & CSP DarkNet53 & PAN & Ultralytics head & - & 97M & 49.4 & 66.3 & 53.6 \\
YOLOv8~\cite{yolov8_ultralytics} & C2f & Enhanced PAN & Ultralytics head & - & 68M & 50.2 & 67.5 & 53.9 \\
YOLOv9~\cite{wang2024yolov9} & CSPNet & PAN & Ultralytics head & - & 58M & 55.2 & 72.2 & 60.3 \\
YOLOv10~\cite{THU-MIGyolov10} & CSPNet & PAN & Ultralytics head & - & 31M & 54.4 & 71.3 & 59.3 \\
YOLOv11~\cite{yolo11_ultralytics} & C3K2 & C2PSA & SPPF & - & 56M & 54.9 & 71.3 & 59.8 \\
\midrule
DiagNet (hard) & DarkNet53 & DiagNet & Ultralytics head & $\mathcal{L}_{\text{min}}$ & 75M & 48.9 & 66.2 & 53.6 \\
 & & & & $\mathcal{L}_{\text{comp}}$ & 75M & 51.5 & 68.7 & 56.1 \\
\midrule
DiagNet (soft) & DarkNet53 & DiagNet & Ultralytics head & $\mathcal{L}_{\text{min}}$ & 75M & 51.4 & 68.6 & 55.6 \\
 & & & & $\mathcal{L}_{\text{comp}}$ & 75M & 53.1 & 71.9 & 58.3 \\
\bottomrule
\end{tabular}
\end{sc}
\end{small}
\vskip 0.0in
\end{table*}

The first experiment was conducted using the Pascal VOC (Visual Object Classes) dataset, a representative benchmark dataset for object detection \cite{everingham2010pascal}. This Pascal VOC dataset consists of 20 object classes, including people, birds, dogs, cats, and cars. The proposed DiagNet was trained using both training data from Pascal VOC 2007 and 2012, mAP$^{50}$ performance which represents the mean average precision calculated at an intersection over union (IoU) threshold of 50 \%, was derived using evaluation data from Pascal VOC 2007 \cite{ren2016faster, redmon2016you}.

The second dataset is the MS COCO (Microsoft Common Objects in Context) dataset, a large-scale dataset consisting of 80 object classes found in real life \cite{lin2014microsoft}. In this paper, like previous studies, we learned a model using the MS COCO dataset released in 2017 and evaluated mAP, mAP$^{50}$, and mAP$^{75}$ performance \cite{yolov5, yolov8_ultralytics}. The mAP refers to mAP$^{50-95}$ which represents the mAP averaged over IoU thresholds from 50 \% to 95 \% in increments of 5 \%. Meanwhile, the mAP$^{75}$ denotes the mAP calculated at a fixed IoU threshold of 75 \%.

\subsection{Implementation details}

\begin{figure*}[ht]
\centering
\includegraphics[width=\textwidth]{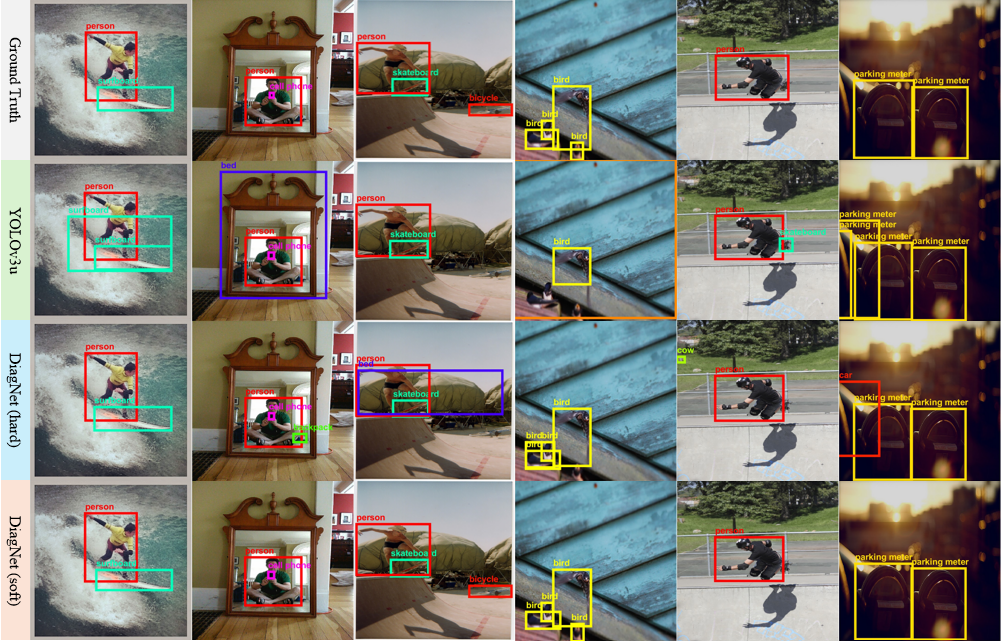}
\captionsetup{skip=3pt}
\caption{Comparisons of detection results on MS COCO dataset.}
\label{icml-historical}
\end{figure*}

We train an object detection model by combining the DiagNet with the detection head of YOLO model. To train and evaluate the DiagNet on the Pascal VOC dataset, we integrate the DiagNet with the detection head of YOLOv1 \cite{redmon2016you}. The backbone network is ResNet-50 \cite{he2016deep}, i.e. pre-trained on ImageNet \cite{deng2009imagenet}. The input of the DiagNet is feature map of size $h=28$ and $c=1024$, extracted by the backbone, and then node embedding and edge prediction using a single-layer GCN are employed. Then the diagonalized map processed with 2D pooling is passed to the detection head. The DiagNet is denoted as two types: DiagNet (hard) trained with (5) and (6), and DiagNet (soft) trained via (9) and (10) with $\alpha = 1$. The object detection model is sequentially trained at each epoch by alternating training of the DiagNet and the YOLOv1 detection head. During training of the YOLOv1 detection head, the learned weights of the DiagNet are fine-tuned.

To train and evaluate the DiagNet on the MS COCO dataset, we integrate the DiagNet with the detection head of YOLOv3u \cite{redmon2018yolov3, yolov5}. For backbone network, we adopt the pre-trained DarkNet53 used in YOLOv3u. The DarkNet53 gives three feature maps of which heights ($h$) are 56, 23, and 13, and the channels ($c$) are 256, 512, and 1024. For handing these feature maps, three separate GCNs are adopted and pyramid pooling is adopted for the DiagNet training. Then, the resulting three diagonalized maps are fed into the detection head. As in the case of the Pascal VOC dataset, the DiagNet (soft) is trained with $\alpha = 1$. The object detection model is trained by alternating training of DiagNet and the YOLOv3u detection head.

\subsection{Detection performance}
We present the mAP$^{50}$ performance of the DiagNet (hard), DiagNet (soft) and YOLOv1 on the Pascal VOC dataset in Table 1. The DiagNet (hard) achieves 1.7 \% higher mAP$^{50}$ than YOLOv1 even it has 44M fewer parameters (\#. Para.). The DiagNet (soft) trained with $L_{\text{comp}}$ achieves the highest mAP$^{50}$ of 58.2 \% which is 7.5 \% higher than that of YOLOv1.

Figure 7 illustrates the detection results of YOLOv1 and DiagNet on the Pascal VOC dataset. Both YOLOv1 and DiagNet detect objects with a detection score threshold of 0.2, and the results of DiagNet are obtained by minimizing $L_{\text{comp}}$. The DiagNet demonstrates superior object detection performance and more accurate bounding box estimation than those of YOLOv1. Note that the DiagNet outperforms YOLOv1 in detecting significantly overlapped multiple objects. Also we can observe that the DiagNet (soft) estimates more accurate bounding boxes and superior detection performance of overlapped objects than those of the DiagNet (hard).

Table 2 shows object detection performance on the MS COCO dataset. The object detection performance is measured by using mAP, mAP$^{50}$, and mAP$^{75}$. The DiagNet (hard) achieves at least 0.9 \% higher mAP than existing models, even it has 28M fewer parameters than YOLOv3u. Note that the DiagNet (soft) trained with $L_{\text{comp}}$ achieves a mAP of 53.1 \%, which is 5.1 \% higher than that of YOLOv3u. Also, we can observe that the DiagNet (soft) outperforms YOLOv5u and YOLOv8 in mAP by at least 2.0 \% and 1.2 \%, respectively. The DiagNet, combined with the backbone and detection head of YOLOv3u, exhibits little lower mAP detection performance than YOLOv9, YOLOv10, and YOLOv11. Note that only 1.3 \% mAP gap to YOLOv9, the model with the highest mAP. In terms of mAP$^{50}$, the DiagNet achieves 0.6 \% higher detection performance than YOLOv10 and YOLOv11. These results implies that the DiagNet has potential to achieve further performance improvements when integrated with different backbones and detection heads. Also, we can observe that the DiagNet outperforms existing two-stage and other one-stage object detectors in terms of mAP.

Figure 8 shows detection results of YOLOv3u and the DiagNet on the MS COCO dataset. Both YOLOv3u and the DiagNet detect objects by using a detection score threshold of 0.2, and results of the DiagNet are obtained by minimizing $L_{\text{comp}}$. As seen in Figure 8, the DiagNet outperforms YOLOv3u in detecting multiple overlapped objects. Also we can observe that the DiagNet (soft) estimates more accurate bounding boxes and superior detection performance of overlapped objects than those of the DiagNet (hard).

\subsection{Training loss over epoch}
We present diagonalization loss and detection loss of the proposed DiagNet on the Pascal VOC dataset when the $L_{\text{comp}}$ is used. Figure 9 presents the convergence of $L_{\text{comp}}$ of the DiagNet and the detection loss of the YOLOv1 detection head over epoch. As shown in Figure 9, we can observe that the $L_{\text{comp}}$ converges to a lower value when it is trained with DiagNet (soft) than with DiagNet (hard) and also the detection loss of the YOLOv1 becomes stable during training.

\begin{figure}[ht]
\vskip 0.0in
\begin{center}
\centerline{\includegraphics[width=\columnwidth]{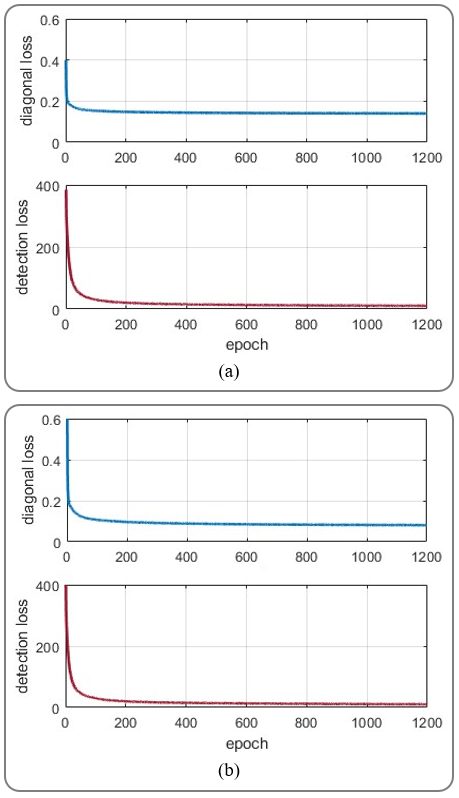}}
\caption{$L_{\text{comp}}$ and detection loss of the YOLOv1 detection head when (a) DiagNet (hard) and (b) DiagNet (soft) are used.}
\label{icml-historical}
\end{center}
\vskip -0.2in
\end{figure}

\begin{figure}[ht]
\vskip 0.0in
\begin{center}
\centerline{\includegraphics[width=\columnwidth]{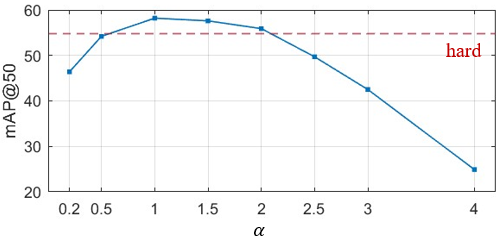}}
\caption{Detection performance according to $\alpha$.}
\label{icml-historical}
\end{center}
\vskip -0.2in
\end{figure}

\subsection{mAP$^{50}$ according to relaxation parameter $\alpha$}
The detection performance of the DiagNet using (9) and (10) depends on the parameter $\alpha$. The change in mAP$^{50}$ detection performance on the Pascal VOC dataset is shown in Figure 10. The DiagNet shows mAP$^{50}$ performance that is at least 1.1 \% higher than the DiagNet with hard constraints when $\alpha$ is between 1 and 2. With $\alpha = 1$, the mAP$^{50}$ is 58.2 \%, which is 3.4 \% higher than that with hard constraint. On the other hand, when $\alpha$ is less than 0.5 or greater than 2, the mAP$^{50}$ performance degrades and becomes lower than result with hard constraint. If $\alpha$ increases over 2, then detection performance significantly drops.

\section{Conclusion}
We have presented the DiagNet, new one-stage detection algorithm based on GCN. By adopting well-known detection head in YOLO models, we showed that the DiagNet using the proposed diagonalization algorithms and loss functions, can improve mAP performance on Pascal VOC and MS COCO dataset.


\bibliography{example_paper}
\bibliographystyle{icml2025}

\newpage
\appendix
\onecolumn
\section{Additional results on Pascal VOC dataset}
Figure 11 presents detection results of the DiagNet (soft) on the Pascal VOC dataset in which we can observe excellent mAP$^{50}$ performance.

\begin{figure}[ht]
\vskip 0.0in
\begin{center}
\centerline{\includegraphics[width=\columnwidth]{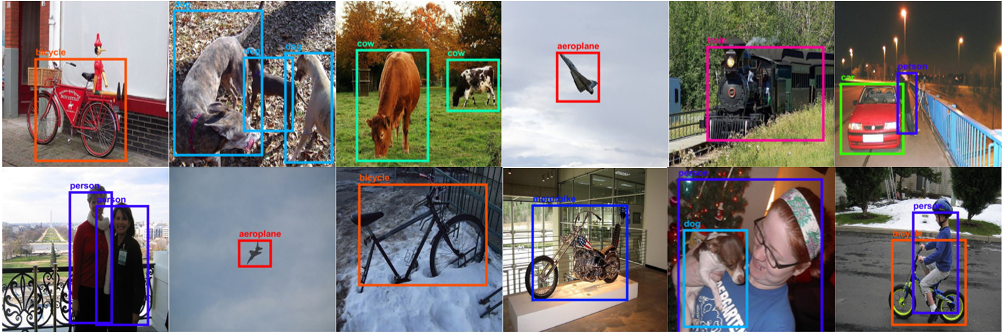}}
\caption{Additional detection results on Pasca VOC dataset.}
\label{icml-historical}
\end{center}
\vskip -0.2in
\end{figure}

\section{Additional results on MS COCO dataset}
Figure 12 presents detection results of the DiagNet (soft) on the MS COCO dataset in which we can observe excellent mAP performance.

\begin{figure}[ht]
\vskip 0.2in
\begin{center}
\centerline{\includegraphics[width=\columnwidth]{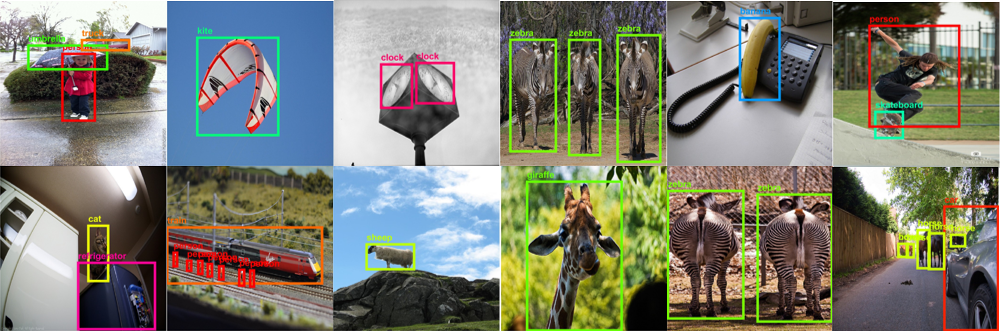}}
\caption{Additional detection results on MS COCO dataset.}
\label{icml-historical}
\end{center}
\vskip -0.2in
\end{figure}

\end{document}